\documentclass{article}

\usepackage{arxiv}                  
\usepackage[utf8]{inputenc}         
\usepackage[T1]{fontenc}            
\usepackage{lipsum}                 
\usepackage{graphicx}               
\graphicspath{{./images/}}
\usepackage{amsmath,amssymb,mathtools}
\usepackage{booktabs}               
\usepackage{multirow}               
\usepackage{tabularx}
\usepackage{enumitem}               
\usepackage{float}                  
\usepackage{subcaption}             
\usepackage{caption}
\usepackage{xcolor}                 
\usepackage{csquotes}               
\usepackage{url}                    
\usepackage[numbers,sort&compress]{natbib} 
\usepackage[ruled,vlined]{algorithm2e}    
\DontPrintSemicolon
\SetAlgoNlRelativeSize{-1}
\SetInd{0.6em}{0.9em}

\usepackage{hyperref}
\hypersetup{colorlinks=true, linkcolor=blue, citecolor=blue, urlcolor=blue}

\usepackage{tikz}
\usetikzlibrary{arrows.meta,positioning,fit,calc}
\usetikzlibrary{backgrounds} 
\usepackage{pgfplots}
\pgfplotsset{compat=1.18}
\usepackage{siunitx}

\setlist{nosep,leftmargin=*}               
\newcommand{\CRS}{\mathrm{CRS}}

\newcommand{\clip}{\operatorname{clip}}

\title{COMPASS: Context-Modulated PID Attention Steering System for Hallucination Mitigation}

\author{
Kenji Sahay \\
Algoverse \\
\texttt{kenjisahay@gmail.com} 
\And
Snigdha Pandya \\
 Algoverse \\
 \texttt{snigdhapandya@gmail.com} 
\And
 Rohan Nagale \\
 Algoverse \\
 \texttt{rohangpu@gmail.com} 
\And
Anna Lin \\
Algoverse \\
\texttt{annalin2121@gmail.com} 
\And
Shikhar Shiromani \\
Georgia Institute of Technology \\
\texttt{ssshiromani3@gatech.edu}
\And
Parham Sharaf \\
University of California, Berkeley \\
\texttt{parham-sharaf@berkeley.edu} 
\And
Kevin Zhu \\
Algoverse \\
\texttt{zhu502846@berkeley.edu} 
\And
Sunishchal Dev \\
Algoverse \\
\texttt{dev@algoverseairesearch.org}
}

\newcommand{\oneC}{\mathbb{1}}
\begin{document}
\vspace*{-0.5cm}

\maketitle
\begin{abstract}

Large language models (LLMs) often produce fluent but factually incorrect statements, even when relevant evidence is available, due to misallocation of attention between contextual inputs and parametric knowledge. Ensuring that models actively reason over context and retrieve relevant information is critical for trustworthy and interpretable AI. We introduce \textbf{COMPASS (Context-Modulated PID Attention Steering System)}, a lightweight, interpretable framework that dynamically steers attention to retrieved context during generation. Using the Context Reliance Score (CRS), COMPASS identifies which attention heads are underutilizing context, and a PID controller adjusts them in real time to improve evidence grounding and factual consistency. This mechanism enables the model to demonstrate advanced reasoning by actively returning to context and retrieving supporting information when needed, without retraining or multi-pass decoding. Across benchmarks including HotpotQA, XSum, HaluEval, and RAGTruth, COMPASS reduces hallucinations by 2.8–5.8\% absolute while revealing how attention heads contribute to context-aligned reasoning. These results show that feedback-driven, interpretable control can enhance reasoning, retrieval, and evidence-based generation in LLMs.

\end{abstract}

\section{Introduction}
LLMs exhibit strong reasoning capabilities but often produce \emph{contextual hallucinations}, where outputs conflict with the input context despite relevant evidence being present \citep{matarazzo2025,zhao2024}. These errors typically arise when the model over-relies on its parametric knowledge or generated history rather than the provided prompt.

Beyond mitigation, COMPASS is designed as a scientific probe of how LLMs use context. Each component,the Context Reliance Score, the classifier, and the PID controller, offers a transparent mapping between internal attention signals and model behavior. Rather than treating interpretability as a post-hoc visualization problem, we embed interpretability in the generation loop itself, allowing real-time observation and modulation of evidence use. This framework provides a principled way to study and steer complex model dynamics.
Existing mitigation strategies include \emph{contrastive or context-aware decoding} \citep{shi2023}, which reweight token probabilities using an auxiliary distribution, and \emph{attention-based diagnostics} such as Lookback Lens \citep{chuang2024}, which train classifiers on attention ratios to detect hallucinations and then guide decoding through candidate re-ranking. More recent approaches, such as DAGCD \citep{huang2025}, intervene directly in the attention mechanism, but often rely on multi-pass decoding or pre-specified head selections, introducing latency and limiting flexibility. Our contributions are four fold:
\begin{itemize}[leftmargin=1.2em]
  \item \textbf{Context-Modulated PID Attention Steering System (COMPASS):} A decoding-time intervention that adjusts attention heads on-the-fly via a \emph{pre-softmax, context-key–only} bias using a real-time diagnostic signal, with no \emph{base-model} retraining or multi-pass decoding.
  \item \textbf{Context Reliance Score (CRS):} The \emph{logit} of the attention mass on \emph{context} keys (last query row), a reformulation of the “lookback ratio” from Lookback Lens \cite{chuang2024}, used as an online per-head context-sensitivity signal for dynamic head selection.
  \item \textbf{Classifier-Guided Conditional Scaling:} Heads are modulated only when a hallucination detector indicates elevated risk, preserving fluency and minimizing unnecessary interventions.
  \item \textbf{Efficient, Interpretable Control:} COMPASS operates within a single decode stream; attentions are read every \(k\) tokens and adjusted via a \emph{pre-softmax, context-key–only} bias, yielding fine-grained, interpretable head-level control.
\end{itemize}

\usetikzlibrary{arrows.meta,positioning,fit,calc}

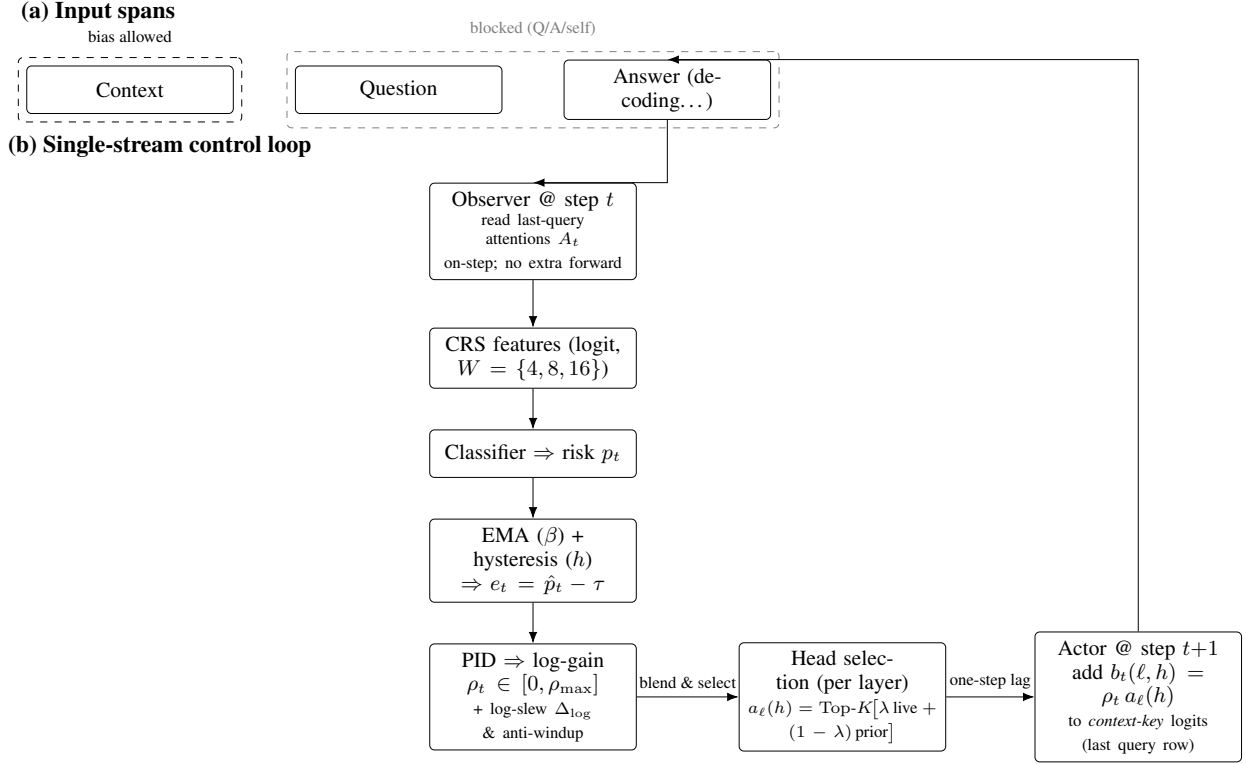
\begin{figure}[t]
  \centering
  \resizebox{\linewidth}{!}{%
\begin{tikzpicture}[
  >=Latex,
  node distance=7mm and 9mm,
  every node/.style={font=\footnotesize},
  box/.style={draw, rounded corners=2pt, align=center, inner sep=1.2mm, fill=white, text width=28mm, minimum height=7mm}
]
  \node[box] (ctx) {Context};
  \node[box, right=9mm of ctx] (q) {Question};
  \node[box, right=9mm of q] (a) {Answer (decoding…)};

  \node[draw, dashed, rounded corners=2pt, fit=(ctx)] (ctxfit) {};
  \node[draw, dashed, rounded corners=2pt, fit=(q)(a), draw=gray] (blocked) {};
  \node[anchor=south, yshift=1mm] at (ctxfit.north) {\scriptsize bias allowed};
  \node[anchor=south, yshift=1mm, text=gray] at (blocked.north) {\scriptsize blocked (Q/A/self)};

  \node[anchor=south west, font=\bfseries] at ([xshift=-2mm,yshift=5mm]ctx.north west) {(a) Input spans};

  \node[box, below=8mm of blocked.south] (attn) {Observer @ step $t$\\\scriptsize read last-query attentions $A_t$\\\scriptsize on-step; no extra forward};
  \draw[->] (a.south) |- (attn.north);

  \node[box, below=7mm of attn] (crs) {CRS features (logit, $W=\{4,8,16\}$)};
  \node[box, below=6mm of crs] (clf) {Classifier $\Rightarrow$ risk $p_t$};
  \node[box, below=6mm of clf] (gate) {EMA ($\beta$) + hysteresis ($h$)\\$\Rightarrow$ $e_t=\hat p_t-\tau$};
  \node[box, below=6mm of gate] (pid) {PID $\Rightarrow$ log-gain $\rho_t \in [0,\rho_{\max}]$\\\scriptsize + log-slew $\Delta_{\log}$ \& anti-windup};

  \draw[->] (attn) -- (crs);
  \draw[->] (crs) -- (clf);
  \draw[->] (clf) -- (gate);
  \draw[->] (gate) -- (pid);

  \node[anchor=south west, font=\bfseries]
    at ($(current bounding box.west|-attn.north)+( -2mm, 2.5mm)$) {(b) Single-stream control loop};

  \node[box, right=15mm of pid] (sel) {Head selection (per layer)\\\scriptsize $a_\ell(h)=\mathrm{Top}\text{-}K\!\big[\lambda\,\text{live}+(1-\lambda)\,\text{prior}\big]$};
  \node[box, right=13mm of sel] (inject) {Actor @ step $t{+}1$\\ add $b_t(\ell,h)=\rho_t\,a_\ell(h)$\\\scriptsize to \emph{context-key} logits (last query row)};

  \draw[->] (pid) -- node[midway, above, font=\scriptsize]{blend \& select} (sel);
  \draw[->] (sel) -- node[midway, above, font=\scriptsize]{one-step lag} (inject);

  \draw[->] (inject.north) |- (a.north);
\end{tikzpicture}%
  }
  \caption{Single-stream control loop and inputs.}
  \label{fig:control-loop}
\end{figure}

\section{Methods}
\label{sec:methods}
\subsection{Problem Setting and Notation}
We study contextual hallucinations: unsupported or factually incorrect tokens given a supplied context. Consider an auto-regressive Transformer \citep{vaswani2017} with $L$ layers and $H$ heads per layer. At generation step $t$, the prompt is partitioned as a fixed \emph{context} $C$ (tokens $1{:}|C|$) followed by a fixed \emph{question} $Q$ (tokens $|C|{+}1{:}|C|{+}|Q|$); the model has produced $t{-}1$ output tokens thereafter. Let $\mathcal{K}_t=\{1,\ldots,|C|{+}|Q|{+}t{-}1\}$ denote key positions, with context keys $\mathcal{K}_C=\{1,\ldots,|C|\}$ and non-context keys $\mathcal{K}_G=\{|C|{+}1,\ldots,|C|{+}|Q|{+}t{-}1\}$ (which include the question and past outputs). For head $h$ in layer $\ell$, let $A_t(\ell,h)\in\mathbb{R}^{|\mathcal{K}_t|}$ be the (causal-masked) attention distribution for the \emph{last query row} at step $t$, and $Z_t(\ell,h)\in\mathbb{R}^{|\mathcal{K}_t|}$ its pre-softmax logits.

Our goal is to bias attention toward the prompt context only when the model is likely to hallucinate, while leaving behavior unchanged when it is grounded. To this end, we design a decode-time intervention that (i) measures each head’s reliance on context vs.\ non-context via a Context Reliance Score (CRS), (ii) predicts token-level hallucination risk from windowed CRS features, and (iii) uses a PID controller to apply a small \emph{pre-softmax, context-key–only} additive bias to selected heads. Formally, when the gate is active we modify
\[
Z_t'(\ell,h,k) \;=\; Z_t(\ell,h,k) \;+\; b_t(\ell,h)\,\mathbf{1}[\,k\in\mathcal{K}_C\,],
\]
on the last-query row only, where $b_t(\ell,h)\!\ge\!0$ is set by the PID output and head selection; all non-context keys and all non-last-query rows remain unchanged.

\subsection{Context Reliance Score (CRS)}
We quantify each head’s context reliance as the fraction of its attention mass on the \emph{prompt-context} keys. For head $(\ell,h)$ at step $t$, with attention distribution $A_t(\ell,h)\in\mathbb{R}^{|\mathcal{K}_t|}$ over all keys in the \emph{last query row} (where $|\mathcal{K}_t|=|C|{+}|Q|{+}t{-}1$), define
\begin{equation}
  p_{\text{ctx}}(t,\ell,h) \;=\; \sum_{i \in \mathcal{K}_C} A_t(\ell,h)[i] \;\in\; [0,1],
\end{equation}
\textit{i.e.}, the total softmax weight on context tokens. For numerical stability and an unbounded signal, we apply a logit transform with clipping:
\begin{align}
  \tilde p_{\mathrm{ctx}}
    &= \clip\!\big(p_{\mathrm{ctx}},\,\varepsilon,\,1-\varepsilon\big),
    & \varepsilon &= 10^{-6}, \label{eq:clip} \\
  \CRS(t,\ell,h)
    &= \log\!\frac{\tilde p_{\mathrm{ctx}}}{1-\tilde p_{\mathrm{ctx}}}.
    \label{eq:crs}
\end{align}
In preprocessing, we compute $\mathrm{CRS}$ for all $(\ell,h)$ across answer time steps and store tensors of shape $[L,H,T_{\text{ans}}]$ \emph{in logit space} (per-head logit of context mass); summary statistics (mean, std., quantiles) are also recorded.

At runtime, we maintain a per-head history \emph{in logit space} (logit of $p_{\text{ctx}}$) and use per-layer $z$-scores to rank heads; this live score can be blended with an optional offline prior.

\paragraph{Feature Extraction.} We compute the Context Reliance Score (CRS) for each head as the fraction of attention mass that the last query places on \emph{prompt-context} keys (i.e., within $\mathcal{K}_C$). For modeling, we apply a logit transform to CRS and compute windowed statistics per head (mean, standard deviation, last-minus-first \emph{delta}) over $W\in\{4,8,16\}$, yielding a feature vector of size $3\cdot |W|\cdot L\cdot H$ (e.g., $9{,}216$ for LLaMA-2-7B). We do not globally standardize these features; runtime head selection uses per-layer $z$-scores of the \emph{live} CRS in probability space, while the classifier consumes the raw windowed features (in logit space).

\subsection{Token-Level Hallucination Risk via Logistic Classifier}
We train a token-level classifier that maps windowed CRS features to a hallucination probability. Our primary model is XGBoost with a logistic objective. Inputs are sliding-window statistics of each head’s recent CRS \emph{logits}: mean, standard deviation, and last-minus-first \emph{delta} which is then computed over $W\in\{4,8,16\}$ (concatenated in increasing $W$) and concatenated across all heads, yielding a feature vector of size $3\cdot|W|\cdot L\cdot H$ (no global standardization).

\begin{table}[ht]
\centering
\caption{Classifier AUROC Performance (Hallucination = Positive Class)}
\label{tab:classifier-auroc}
\begin{tabular}{l l S[table-format=1.3]}
\toprule
\textbf{Model} & \textbf{Dataset} & {\textbf{AUROC}} \\
\midrule
\multirow{4}{*}{Qwen-2.5-7B-Instruct} 
    & HotpotQA   & 0.839 \\
    & XSum       & 0.953 \\
    & RAGTruth   & 0.789 \\
    & HaluEval   & 0.886 \\
\midrule
LLaMA-2-7B      & RAGTruth & 0.858 \\
LLaMA-2-13B     & RAGTruth & 0.873 \\
Mistral-7B      & RAGTruth & 0.912 \\
\bottomrule
\end{tabular}
\end{table}
\paragraph{Classifier Training.}
We use \textbf{XGBoost} with a logistic objective to map windowed CRS features to hallucination risk. XGBoost handles nonlinear interactions among heads/layers, runs efficiently for repeated decode-time queries, and provides per-feature importances that we aggregate into per-(layer, head) weights. These weights serve as a \emph{static prior} for the online modulator and are \emph{blended} with live per-layer $z$-scores during head selection. Data is split 70/10/20 into Train/Validation/Test by example id to prevent leakage across partitions.

\textbf{Runtime use.} The online modulator constructs the same windowed feature vector from live CRS histories (for all $W$) and queries the classifier every $k$ tokens to obtain $p_t\in[0,1]$, which feeds the EMA+hysteresis-gated PID loop.

\resizebox{\linewidth}{!}{%
\begin{tikzpicture}[>=Latex, font=\small, node distance=6mm and 10mm]
  \tikzset{
    block/.style={draw, rounded corners, fill=white, minimum width=16mm, minimum height=7mm, align=center, inner sep=2pt},
    sum/.style={draw, circle, fill=white, minimum size=6mm, inner sep=0pt},
    sig/.style={midway, fill=white, inner sep=1pt}
  }

  \node (rin)  at (-1.8,  0.8) {$r^{t}$};
  \node (rset) at (-1.8, -0.8) {$r^{\ast}$};

  \node[sum]   (sumE)  at (0,0) {$+$};
  \coordinate  (split) at (0.8,0);

  \node[block, above=8mm of split] (P) {$K_{p}$\\ \scriptsize proportional};
  \node[block, right=12mm of split] (I) {$K_{i}$\\ \scriptsize integrator};
  \node[block, below=8mm of split] (D) {$K_{d}$\\ \scriptsize differentiator};

  \node[sum,  right=12mm of I] (sumU) {$+$};
  \node[block, right=10mm of sumU] (EMA) {EMA $(\beta)$\\[-1pt]\scriptsize $y^{t}{=}\beta y^{t-1}{+}(1{-}\beta)u^{t}$};
  \node[block, right=10mm of EMA]  (Gate) {Hysteresis\\[-1pt]\scriptsize $r^{t}>\tau_{\text{on}}$};
  \node[block, right=10mm of Gate] (Slew) {Slew limit\\[-1pt]\scriptsize $\Delta\log\rho \le \Delta_{\max}$};
  \node[block, right=10mm of Slew] (Sat)  {Saturation\\[-1pt]\scriptsize $[0,\rho_{\max}]$};
  \node[right=8mm of Sat] (out) {$\rho^{t}$};

  \begin{scope}[on background layer]
    \draw[->] (rin)  -- ($(sumE.west)+(0,0.4)$);
    \draw[->] (rset) -- node[sig] {$-$} ($(sumE.west)+(0,-0.4)$);

    \draw[->] (sumE) -- node[sig] {$e^{t}$} (split);

    \draw[->] (split) |- (P);
    \draw[->] (split) --  (I);
    \draw[->] (split) |- (D);

    \draw[->] (P) -| node[sig] {$P$} (sumU);
    \draw[->] (I) -- node[sig] {$I$} (sumU);
    \draw[->] (D) -| node[sig] {$D$} (sumU);

    \draw[->] (sumU) -- node[sig] {$u^{t}$} (EMA);
    \draw[->] (EMA) -- (Gate);
    \draw[->] (Gate) -- (Slew);
    \draw[->] (Slew) -- (Sat);
    \draw[->] (Sat)  -- (out);

    \draw[densely dotted, ->]
      ($(Sat.south west)!0.5!(Sat.south)$) |- ++(-0.2,-0.8) -| (I.south);
  \end{scope}

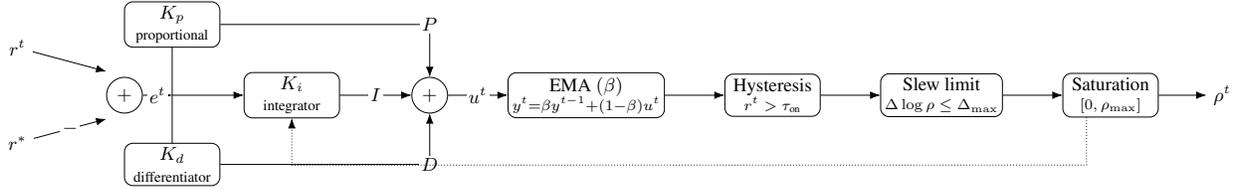
  \node[align=left, anchor=west] at ($(rin)+(-0.1,1.2)$) {\bfseries Figure 2: PID with EMA, hysteresis, and limits.};
  
\end{tikzpicture}
}

\subsection{Head Selection and Scaling}
Algorithms~\ref{alg:heads}–\ref{alg:pid} summarize the controller and per-step head modulation. In brief, we rank heads within a mid-to-upper layer range by blending per-layer $z$-scored live CRS with a static prior $w(\ell,h)$, keep the top-$K$ per layer (set via \texttt{--keep-per-layer}; $K{=}16$ in our runs), renormalize the weights, and add a \emph{pre-softmax, context-only} bias of magnitude $\rho_t\,a_\ell(h)$ to the last-query row. Non-context keys and all non-last-query rows remain unchanged
\begin{algorithm}[t!]
  \caption{Head Selection \& Context-Key Bias (per risk step)}
  \label{alg:heads}
  \footnotesize
  \SetKwInOut{KwParams}{Params}
  \SetKwInOut{KwInput}{Input}
  \SetKwInOut{KwOutput}{Output}

  \KwParams{layer range $[\lfloor L/2\rfloor,\,L)$;\ keep per layer $K$;\ blend $\lambda$}
  \KwInput{on-step attentions $A_t$ (\texttt{output\_attentions=true}); prior $w(\ell,h)$; gain $\rho_t$}
  \KwOutput{pre-softmax bias on \emph{context keys} (last query row)}

  \If{$\rho_t=0$}{\textbf{return} (no intervention)}
  \For{$\ell \in [\lfloor L/2\rfloor,\,L)$}{
    compute live CRS logits $v_\ell(h)$ from $A_t$\;
    $z_\ell(h) \leftarrow \max\{0,\,\text{zscore}_{\text{per-layer}}(v_\ell(h))\}$\;
    $\tilde a_\ell(h) \leftarrow \lambda\,\mathrm{norm}_{[0,1]}(z_\ell(h)) + (1-\lambda)\,w(\ell,h)$\;
    $S_\ell \leftarrow \mathrm{Top}\text{-}K\_h\,\tilde a_\ell(h)$;\quad
    $a_\ell(h) \leftarrow \tilde a_\ell(h)\big/\sum_{h\in S_\ell}\tilde a_\ell(h)$\;
    \For{$h \in S_\ell$}{
      $Z_t(\ell,h)[1{:}|C|] \leftarrow Z_t(\ell,h)[1{:}|C|] + \rho_t\,a_\ell(h)$ \tcp*[r]{context keys, last query row}
    }
  }
  \tcc{$z(\cdot)$ is per-layer $z$-score; $\mathrm{norm}_{[0,1]}$ rescales over heads in a layer.}
\end{algorithm}

\begin{algorithm}[t!]
  \caption{PID–Gated Log–Gain (Controller)}
  \label{alg:pid}
  \footnotesize
  \SetKwInOut{KwParams}{Params}
  \SetKwInOut{KwInput}{Input}
  \SetKwInOut{KwOutput}{Output}

  \KwParams{$\tau$ (target), $h$ (hysteresis), $\beta$ (EMA), $(K_p,K_i,K_d)$,\\
            $\rho_{\max}$ (cap), $\Delta_{\log}$ (log-slew), $\varepsilon$ (small)}
  \KwInput{$(\hat p_{t-1},I_{t-1},\rho_{t-1})$, new risk $p_t$}
  \KwOutput{$(\hat p_t,I_t,\rho_t)$}

  $\hat p_t \leftarrow \beta\,\hat p_{t-1} + (1-\beta)\,p_t$\;
  $e_t \leftarrow \hat p_t - \tau$;\quad \If{$|e_t|\le h$}{$e_t \leftarrow 0$}
  $P \leftarrow K_p e_t$\;
  \uIf{$(\rho_{t-1}=0 \land e_t<0)\ \lor\ (\rho_{t-1}=\rho_{\max} \land e_t>0)$}{$I \leftarrow I_{t-1}$}
  \Else{$I \leftarrow I_{t-1} + K_i e_t$}
  $D \leftarrow K_d(\hat p_t-\hat p_{t-1})$\;
  $\rho_{\mathrm{raw}} \leftarrow \mathrm{clip}(P+I+D,\,0,\,\rho_{\max})$\;
  $\ell_{\mathrm{prev}} \leftarrow \log(\rho_{t-1}+\varepsilon)$;\quad
  $\ell_{\mathrm{raw}} \leftarrow \log(\rho_{\mathrm{raw}}+\varepsilon)$\;
  $\ell \leftarrow \ell_{\mathrm{prev}} + \mathrm{clip}(\ell_{\mathrm{raw}}-\ell_{\mathrm{prev}},-\Delta_{\log},\Delta_{\log})$\;
  $\rho_t \leftarrow e^{\ell}-\varepsilon$;\quad $I_t \leftarrow I$\;
  \tcc{Outputs nonnegative log–gain $\rho_t$ with anti-windup and slew limiting.}
\end{algorithm}
\subsection{Pre-Softmax Attention Bias}
We add a context-only, last-query-row bias to attention logits. For each selected head $(\ell,h)$ and context index $i\in C$,
\[
\tilde Z_t(\ell,h)[i] = Z_t(\ell,h)[i] + b_t(\ell,h), \qquad b_t(\ell,h)\doteq \rho_t\,a_\ell(h),
\]
while non-context keys remain unchanged, $\tilde Z_t(\ell,h)[j] = Z_t(\ell,h)[j]$ for $j\notin C$. The updated attention is $\tilde A_t(\ell,h) = \operatorname{softmax}(\tilde Z_t(\ell,h))$.  

Adding the bias in logit space multiplies the affected unnormalized weights by $\exp(b_t(\ell,h))$, preserving softmax normalization:
\begin{align}
  \exp\!\{\tilde Z_t(\ell,h)[i]\}
    &= e^{\,b_t(\ell,h)}\,\exp\!\{Z_t(\ell,h)[i]\}, && i\in C, \label{eq:psb-mult}\\
  \alpha_t(\ell,h) &\doteq \exp\!\big(\rho_t\,a_{\ell}(h)\big), \label{eq:psb-alpha}\\
  \tilde Z_t &= Z_t + \log \alpha_t \cdot \oneC . \label{eq:psb-compact}
\end{align}

We reset the bias each step and only apply it when the controller is active; non-context keys and all non-last-query rows are never modified.

\subsection{Full Decoding-Time Algorithm}

\paragraph{Inputs.}
Model $f$; prompt context $C$; detector $f_{\text{det}}$ (XGBoost) with threshold $\tau$; hysteresis $h$; PID gains $(k_P,k_I,k_D)$; layer subset $\mathcal{L}^\star$; update cadence $k$ (compute risk/selection every $k$ tokens); window set $W$ for features (default $W=\{4,8,16\}$); per-layer head budget $K$ (default $K=16$); prior–blend weight $\lambda$ (default $\lambda=0.3$); log-space slew limit $\Delta_{\log}$ (default $0.20$); log-gain cap $\rho_{\max}$ (default $1.0$).

\paragraph{Per-step loop for $t=1,2,\dots$}
\begin{enumerate}[leftmargin=1.4em]
  \item \textbf{Attention read (every $k$ tokens).} When $t \bmod k = 0$, enable \texttt{output\_attentions} and read the \emph{last-token query row} per head \emph{on the same forward pass} (no extra forward). Otherwise, reuse the last risk and head selection.
  \item \textbf{CRS \& features.} From that last-query row, compute $\CRS_t(\ell,h)$ as the fraction of attention mass on context keys, then form sliding-window features in the \emph{logit} domain (mean, std, and end–minus–start trend) for $W\in\{4,8,16\}$; concatenate across windows.
  \item \textbf{Risk prediction.} Feed features to $f_{\text{det}}$ to obtain $p_t$; apply EMA smoothing to get $\hat p_t$ and a dead-band $|\hat p_t-\tau|\le h$ (hysteresis).
  \item \textbf{PID update (nonnegative log-gain).} If outside the dead-band, update $(P,I,D)$ on the error $e_t=\hat p_t-\tau$ and produce a nonnegative log-gain $\rho_t$. Apply a \emph{log-space} slew limit with step size $\Delta_{\log}$ and clamp to $[0,\rho_{\max}]$; set $\alpha^t=\exp(\rho_t)$.
  \item \textbf{Head selection.} For each $\ell\in\mathcal{L}^\star$, $z$-score the live head vector per layer, clamp negatives to $0$, min–max normalize to $[0,1]$, blend with the prior via $a=\lambda\,\text{live}+(1-\lambda)\,\text{prior}$, and keep the top-$K$ heads.
  \item \textbf{One-step-lag actuation (pre-softmax).} At step $t{+}1$, add the bias $\log \alpha^t$ to the \emph{context-key} logits of the selected heads on the last-query row only; non-context keys and all non-last-query rows remain unchanged.

\paragraph{Complexity.}
COMPASS reuses attention tensors produced on the same decode step when \texttt{output\_attentions} is enabled (every $k$ tokens); no extra forward pass is introduced. The additional work per risk step is: computing CRS from the read attentions, a small set of vector ops for windowed features, a lightweight classifier call, and a few scalar updates for the PID and log-slew. The only tensor write is adding the pre-softmax bias at selected context indices. Empirically the overhead is modest on 7B models, and decoding remains a single stream.

\end{enumerate}

\section{Risk Calculation Details}
For clarity, we summarize the end-to-end loop executed at \emph{risk steps} ($t \bmod k = 0$):
\begin{enumerate}[leftmargin=1.4em]
  \item \textbf{Attention read \& CRS:} On the same forward pass, enable \texttt{output\_attentions} and read the \emph{last-query row}. For each $\ell\in\mathcal{L}^\star$ and head $h$, compute the context mass on keys $1{:}|C|$ and its logit to obtain $\CRS_t(\ell,h)$; form sliding-window CRS-\emph{logit} features (mean, std, end--minus--start) over $W\in\{4,8,16\}$ and concatenate across windows (no dataset-wide standardization).
  \item \textbf{Predict risk:} Feed the windowed features to $f_{\text{det}}$ to obtain $p_t\in[0,1]$ and compute the smoothed score $\hat p_t$ via EMA; apply a hysteresis dead-band $|\hat p_t-\tau|\le h$.
  \item \textbf{PID update (nonnegative log-gain):} If outside the dead-band, update $(P,I,D)$ on $e_t=\hat p_t-\tau$ with anti-windup, producing a nonnegative log-gain $\rho_t$; apply a log-space slew limit $\Delta_{\log}$ and clamp to $[0,\rho_{\max}]$.
  \item \textbf{Head selection:} For each layer, $z$-score the live CRS vector per layer, clamp negatives to $0$, min--max normalize to $[0,1]$, blend with the static prior $w(\ell,h)$ via $a=\lambda\,\text{live}+(1-\lambda)\,\text{prior}$, select the top-$K$ heads, and renormalize to weights $a_\ell(h)$.
  \item \textbf{Actuation (pre-softmax, one-step lag):} At step $t{+}1$, add a context-only bias of magnitude $\rho_t\,a_\ell(h)$ to the selected heads’ logits on the last-query row: $Z_{t+1}(\ell,h)[1{:}|C|]\leftarrow Z_{t+1}(\ell,h)[1{:}|C|]+\rho_t\,a_\ell(h)$. Non-context keys and all non--last-query rows remain unchanged.
  \item \textbf{Generate token:} Finish the forward pass and sample the next token as usual.

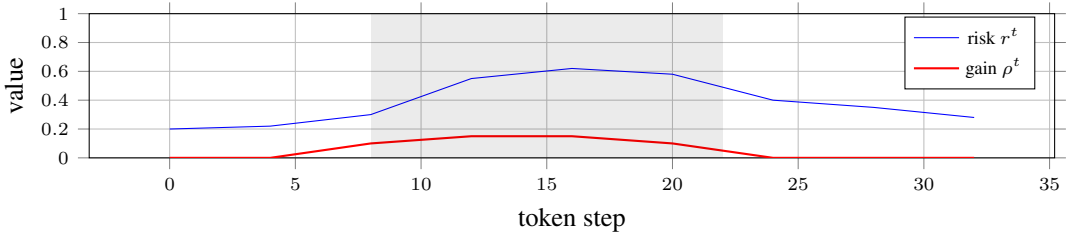
\begin{tikzpicture}
\begin{axis}[
  width=0.9\linewidth, height=3.5cm,
  xlabel={token step}, ylabel={value},
  ymin=0, ymax=1,
  legend style={at={(0.98,0.98)},anchor=north east,font=\scriptsize},
  grid=both, tick align=outside,
  tick label style={font=\scriptsize}
]
  \addplot+[mark=none] table[row sep=\\] {
  x y\\ 0 0.20\\ 4 0.22\\ 8 0.30\\ 12 0.55\\ 16 0.62\\ 20 0.58\\ 24 0.40\\ 28 0.35\\ 32 0.28\\};
  \addlegendentry{risk $r^t$}

  \addplot+[mark=none, thick] table[row sep=\\] {
  x y\\ 0 0.00\\ 4 0.00\\ 8 0.10\\ 12 0.15\\ 16 0.15\\ 20 0.10\\ 24 0.00\\ 28 0.00\\ 32 0.00\\};
  \addlegendentry{gain $\rho^t$}

  \addplot [draw=none, fill=black, fill opacity=0.08] coordinates {(8,0) (22,0) (22,1) (8,1)};
\end{axis}
\end{tikzpicture}
\end{enumerate}

This procedure is well-posed: CRS is bounded and interpretable; the detector outputs calibrated probabilities under a logistic objective; head influence is transparent via $a_\ell(h)$; and the PID (with EMA, hysteresis, and log-space slew) responds to sustained risk while suppressing transient noise.

\paragraph{Hyperparameters.}
Unless noted, the classification threshold $\tau$ is tuned on a dev split (maximize F$_1$) and hysteresis width is $h=0.01$. PID gains: $k_P=0.8$, $k_I=0.2$, $k_D=0.0$; EMA $\beta=0.8$; update cadence $k=1$ (compute risk/selection every token by default); head budget $K=16$ per layer; windows $W=\{4,8,16\}$; prior blend $\lambda=0.3$; log-space slew limit $\Delta_{\log}=0.20$; log-gain cap $\rho_{\max}=1.0$. By default we act on mid-to-upper layers (upper half of the stack); a different subset $\mathcal{L}^\star$ can be provided via configuration.
\paragraph{Ablations.}
We ablate: (i) no PID (threshold+gate only), (ii) no classifier (heuristic CRS-based risk), (iii) layer range choices (last layer only vs.\ mid-to-upper vs.\ all layers), (iv) keep-per-layer $K\in\{4,8,16,32\}$ and prior blend $\lambda\in[0,1]$, (v) log-gain parameters $\rho_{\max}$ and $\Delta_{\log}$, and (vi) update cadence $k\in\{1,2,4\}$. 
\paragraph{Latency.}
COMPASS reuses attentions produced on the same decode step whenever \texttt{output\_attentions} is enabled (every $k$ tokens); no extra forward pass is introduced. Per risk step, the additional work is: computing CRS from the read attentions, a small set of vector ops to form windowed features, a lightweight classifier call, and a few scalar PID/log-slew updates. The only tensor write is adding a pre-softmax bias at selected context indices. 

\section{Experimental Setup}

Our experiments proceed in two stages. Phase~1 constructs a high-precision hallucination detector that operates during decoding using only attention-based features. Phase 2 integrates this detector into the generation loop and dynamically rescales attention heads that are automatically selected at runtime as context-reliant using the Context Reliance Score (CRS), applying modulation only when hallucination risk is high. We obtain the needed attention signals with occasional auxiliary reads every $k$ tokens (via \texttt{--risk-step}), minimizing overhead. Together, these stages test whether lightweight, real-time control of internal attention can improve factual faithfulness without training or multi-pass decoding.

\subsection{Phase 1: Detector Construction and Validation}
\paragraph{Data \& Labeling.}
We generate answers with LLaMA-2-7B, LLaMA-2-13B, Mistral-7B-Instruct, and Qwen-2.5 on four faithfulness-critical benchmarks: HotpotQA (open-domain QA), XSum (single-sentence summarization), HaluEval (hallucination evaluation in QA and summarization), and RAGTruth (adversarial fact-checking). Each answer is split into span-level substrings. An automatic verifier (Gemini~2.5-Flash, structured JSON schema) labels each example with \texttt{is\_hallucination} and up to $0$--$5$ \texttt{unsupported\_spans}, plus a brief analysis and confidence; adjudication uses the same model at temperature~$0.0$. We manually annotated a random sample of $100$ spans and found a $93\%$ agreement between Gemini~2.5-Flash's annotations and human judgments, confirming high consistency.

\subsection{Phase 2: Modulating Attention Heads}
\paragraph{Head Selection.} At runtime we compute a blended score per head:
\begin{equation}
    s_t(\ell,h)=\lambda\, z\!\operatorname{CRS}_t(\ell,h)+(1-\lambda)\,\text{prior}(\ell,h),
\end{equation}
where the live term uses per-layer $z$-scores and the prior comes from exported head importance. We then select the top-$k$ heads per layer over a default mid-to-upper range (layers $16$–$31$ in a 32-layer LLaMA-2-7B), configurable via \texttt{--layers}.\\
\textbf{Control Loop.}
Every \(k\) tokens we read on-step attentions (\texttt{output\_attentions=true}; no extra forward), compute a token-level risk \(p_t\) (EMA-smoothed) with hysteresis gating, and pass it to a PID controller to produce an intensity \(\rho_t \in [0,\rho_{\max}]\). We then add a pre-softmax bias for the \emph{context-key} logits of the selected heads at the last query row (equivalent to \(\log\)-gains), leaving non-context keys untouched.\\
\textbf{Models and Baselines.} Modulation is evaluated on LLaMA-2~7B and Mistral~7B. Baselines include (i) identical sampling path with mitigation disabled, (ii) Lookback Lens guided re-ranking, (iii) Contrastive Decoding \citep{li2023}, and (iv) random-head scaling (same $\alpha$ but on a random head subset).

\section{Results}
\begin{table}[ht]
\centering
\caption{ Results (Hallucination Reduction)}
\label{tab:results} 
\begin{tabular}{@{} l l c c c @{}}
\toprule
\textbf{Model} & \textbf{Dataset} & \textbf{Mitigation Rate $\downarrow$} & \textbf{Span Density $\downarrow$} & \textbf{Context Overlap $\uparrow$}\\
\midrule
\multirow{4}{*}{Qwen-2.5-7B-Instruct} 
    & HotpotQA   & 4.2\% & -14.2\% & +0.06\\
    & XSum       & \textbf{2.8\%} & -11.4\% & +0.04\\
    & RAGTruth   & 3.1\% & \textbf{-16.7\%} & \textbf{+0.08}\\
    & HaluEval   & 5.8\% & -13.8\% & +0.05\\
\midrule
LLaMA-2-7B      & RAGTruth & \textbf{4.2\%} & -18.3\% & +0.09\\
LLaMA-2-13B     & RAGTruth & 5.8\% & \textbf{-22.4\%} & \textbf{+0.12}\\
Mistral-7B      & RAGTruth & 4.9\% & -20.1\% & +0.11\\
\bottomrule
\end{tabular}
\end{table}

 We evaluated COMPASS, our Context-Modulated PID Attention Steering System, on LLaMA-2-7B, LLaMA-2-13B, Mistral-7B, and Qwen-2.5-7B across four benchmarks probing different aspects of contextual hallucination: RAGTruth, HotpotQA, XSum, and HaluEval. We evaluate hallucination reduction using three complementary metrics to show that reductions in hallucination come from better grounding rather than blunt suppression:

\begin{itemize}[leftmargin=1.2em]
  \item \textbf{Mitigation Rate (MR):} The absolute reduction in hallucination rate compared to the unmodified baseline model.
  
  \item \textbf{Span Density (SD):} The number of unsupported spans per 100 generated tokens. A span is considered \emph{unsupported} if it lacks a 3–5-gram match in the provided context and fails the sentence-level factual verifier.
  
  \item \textbf{Context Overlap (CO):} The fraction of generated tokens whose aligned \(n\)-grams appear in the retrieved context, serving as a proxy for grounding distinct from surface-level repetition.
\end{itemize}

These results indicate that COMPASS consistently reduces hallucination rates relative to unmodified baselines, with absolute reductions ranging from approximately 2.8\% to 5.8\% depending on the model and dataset. For instance, on RAGTruth, LLaMA-2-13B achieved a 5.8\% reduction in hallucination rate, while LLaMA-2-7B saw a 4.2\% decrease. HotpotQA and XSum also showed improvements in multi-hop reasoning accuracy and summarization faithfulness, respectively. Span density of unsupported content decreased across all datasets. Context overlap (CO)increased or remained stable suggesting that COMPASS preserves model grounding without excessively perturbing attention. Across the board, larger models (e.g., LLaMA-2-13B, Mistral-7B) benefited more from attention modulation, likely reflecting richer redundancy and more exploitable head-level structure. These findings support the feasibility of lightweight, real-time attention modulation for mitigating contextual hallucinations without multi-pass decoding or retraining. To keep comparisons compute-fair while covering diverse failure modes, we evaluate Qwen-2.5-7B across all four datasets and use RAGTruth as a shared grounded-QA setting for the LLaMA-2 and Mistral models.

\section{Discussion}
Lookback Lens detects and mitigates contextual hallucinations by monitoring the “lookback ratio” (attention to source context vs. newly generated tokens) and guiding decoding with a lightweight classifier; it transfers across tasks/models and reports measurable reductions (e.g., ~9.6\% on XSum) without model retraining. By contrast, COMPASS preserves the full prompt and steers head-level attention using a PID controller keyed to an online context-reliance signal. Empirically (table~\ref{tab:classifier-auroc}), COMPASS achieves lower hallucination rates alongside higher CO and lower SD, indicating that steering internal attention during generation is competitive with (and complementary to) Lookback Lens’s classifier-guided decoding approach. Compared with other approaches, decoding-only tweaks (temperature/nucleus/repetition) do not explicitly target evidence alignment and yield smaller or inconsistent mitigation; self-consistency can help but multiplies decoding cost; static head ablations capture some benefit but cannot adapt to example-specific evidence patterns. COMPASS achieves single-pass mitigation via per-token head modulation, which we observe as reduced SD without sacrificing CO.

\section{Limitations}
While our dynamic attention head modulation framework reduces hallucinations, several constraints remain. First, head-importance estimation is driven by short-horizon, per-step last-query attention signals and may under-perform in extremely long-context or multi-turn settings where risk accrues gradually without strong local cues. Second, gating decisions are made per step without global discourse awareness, so suppression of longer narrative drifts is limited. Third, our context-modulated PID attention steering system introduces sensitivity to controller hyperparameters: EMA smoothing and hysteresis (\(\beta, h\)), PID gains \((k_P,k_I,k_D)\), and the log-space slew limit and cap \((\Delta_{\log}, \rho_{\max})\); poor tuning can cause oscillatory activation, unstable convergence, or oversuppression of useful heads. Finally, although decoding remains single-stream and attentions are read on-step every \(k\) tokens (no second forward pass), computing CRS features and per-layer top-\(K\) selection adds modest but non-negligible overhead that can impact latency on smaller GPUs or very long sequences. Our approach also assumes that head importance can be estimated reliably in real time from live CRS (optionally blended with a prior), an assumption that may degrade in highly non-stationary domains. Evaluation to date focuses on standard benchmarks; open-world and adversarial settings remain to be tested.

\section{Related Work}
LLMs frequently generate fluent but factually incorrect content (``hallucination''). Prior strategies include grounding with retrieval \citep{lewis2009}, structured knowledge graphs, or external consistency classifiers \citep{zhao2024}. Reinforcement learning from human feedback \citep{christiano2023} further improves reliability \citep{ouyang2022}. However, most approaches act post hoc, correcting outputs after hallucinations emerge, rather than intervening in the model's internal reasoning. Transformer heads vary widely in function and importance \citep{voita2019}. Some heads are strongly tied to factual grounding, while others promote hallucinations \citep{chuang2024} Work on pruning and masking \citep{zhang2022,li2025} shows selective head control can shift model behavior, but interventions are static. Our method instead uses dynamic, classifier-informed modulation, adjusting hallucination-prone heads online during decoding. Control theory, particularly PID feedback, has been widely applied in dynamical systems \citep{astrom2008feedback}
, optimization 
\citep{Lin2021} , and reinforcement learning \citep{lewis2009}. We frame hallucination control as a feedback problem: a classifier monitors drift, while a PID loop gates attention heads in real time. Unlike retrieval-based grounding \citep{shuster2021}\citep{lewis2009} \citep{christiano2023}~\citep{ouyang2022}
, or static head pruning \citep{voita2019, zhang2022}, our contribution is the \emph{first closed-loop framework for hallucination mitigation that unifies detection and internal modulation via PID control}.

\section{Conclusion}

COMPASS introduces a lightweight, interpretable, and real-time approach to mitigating contextual hallucinations in LLMs. By embedding a PID-controlled feedback loop into the decoding process and leveraging the Context Reliance Score as a per-head grounding signal, the system achieves preliminary reductions of 2.8--5.8\% in hallucination rate, decreases unsupported-span density, and improves context overlap metrics without retraining or multi-pass decoding. The results suggest that attention-level control can complement traditional post hoc mitigation methods, providing a fine-grained mechanism for enhancing factual accuracy during generation.

Mathematically, COMPASS demonstrates that closed-loop feedback applied to attention logits, modeled as:
\[
\tilde{Z}_t(\ell, h)[i] = Z_t(\ell, h)[i] + \rho_t a_\ell(h), \quad i \in C,
\]
can dynamically steer model outputs toward contextually supported tokens while leaving other attention weights unchanged. This approach highlights the potential for control-theoretic methods in LLM alignment, offering an interpretable, modular alternative to more opaque interventions like fine-tuning or contrastive decoding.

Future work will explore: (i) the integration of richer detector signals that capture semantic coherence, and (ii) formalizing stability guarantees for PID-controlled attention modulation. Overall, these preliminary results validate the feasibility of feedback-driven attention control as a scalable, low-overhead strategy for reducing hallucinations in modern LLMs.

\section{Ethics Statement}
This study makes exclusive use of open-source datasets and pre-existing model checkpoints; no personally identifiable information was collected or processed. All resources were accessed under their respective licenses and applied only for research purposes. Our approach aims to strengthen factual grounding in language models, with potential benefits for downstream systems that rely on trustworthy text generation. Nonetheless, inherent risks of bias, misinformation, or offensive outputs remain, underscoring the need for careful monitoring and responsible deployment.

\bibliographystyle{plain}
\bibliography{references}

\end{document}